\documentclass[11pt]{article}
\usepackage{microtype}

\usepackage{amsmath,amssymb}
\usepackage{amsfonts,bm,bbm} 
\usepackage{amssymb,amsthm,enumitem}
\usepackage{tikz}
\usetikzlibrary{decorations.pathmorphing,positioning,patterns,snakes}
\usepackage{multirow}
\usepackage{xcolor,graphicx,float,booktabs}
\usepackage{tabularx}
\newcolumntype{L}[1]{>{\raggedright\arraybackslash}m{#1}}
\newcolumntype{C}[1]{>{\centering\arraybackslash}m{#1}}
\newcolumntype{R}[1]{>{\raggedleft\arraybackslash}m{#1}}

\usepackage[algo2e,ruled	,vlined]{algorithm2e}

\usepackage{epstopdf}

\usepackage{verbatim} 
\allowdisplaybreaks[4]
\usepackage{authblk}

\usepackage{hyperref}
\hypersetup{colorlinks=true,
linkcolor=blue,
anchorcolor=blue,
citecolor=blue}

\usepackage[round]{natbib}
\usepackage[
margin=1in,
includefoot,
footskip=30pt,
]{geometry}

\newcommand{\bx}{\textbf{x}}

\newcommand{\cX}{\mathcal{X}}
\newcommand{\E}{\mathbb{E}}
\newcommand{\R}{\mathbb{R}}
\newcommand{\Var}{\mathrm{Var}}
\newcommand{\MSE}{\mathrm{MSE}}

\newtheorem{theorem}{Theorem}[section]

\numberwithin{equation}{section}

\title{Derivative-Free Optimization via Finite Difference Approximation: An Experimental Study}

\author[1, 2]{Du-Yi Wang\footnote{tylzml@ruc.edu.cn}}
\author[1]{Guo Liang\footnote{liangguo000221@ruc.edu.cn}}
\author[2]{Guangwu Liu\footnote{msgw.liu@cityu.edu.hk}}
\author[1]{Kun Zhang\footnote{kunzhang@ruc.edu.cn}}
\affil[1]{Institute of Statistics and Big Data\protect\\Renmin University of China\protect\\Beijing, China}
\affil[2]{Department of Decision Analytics and Operations\protect\\City University of Hong Kong\protect\\Tat Chee Avenue, Kowloon, Hong Kong, China}

\begin{document}
\normalsize

\maketitle

\begin{abstract}

Derivative-free optimization (DFO) is vital in solving complex optimization problems where only noisy function evaluations are available through an oracle. Within this domain, DFO via finite difference (FD) approximation has emerged as a powerful method. Two classical approaches are the Kiefer-Wolfowitz (KW) and simultaneous perturbation stochastic approximation (SPSA) algorithms, which estimate gradients using just two samples in each iteration to conserve samples. However, this approach yields imprecise gradient estimators, necessitating diminishing step sizes to ensure convergence, often resulting in slow optimization progress. In contrast, FD estimators constructed from batch samples approximate gradients more accurately. While gradient descent algorithms using batch-based FD estimators achieve more precise results in each iteration, they require more samples and permit fewer iterations. This raises a fundamental question: which approach is more effective—KW-style methods or DFO with batch-based FD estimators? This paper conducts a comprehensive experimental comparison among these approaches, examining the fundamental trade-off between gradient estimation accuracy and iteration steps. Through extensive experiments in both low- and high-dimensional settings, as well as hyperparameter optimization adjusting the penalty parameter in ridge regression, we demonstrate a {\it surprising} finding: when an efficient batch-based FD estimator is applied, its corresponding gradient descent algorithm generally shows better performance compared to classical KW and SPSA algorithms in our tested scenarios.

\end{abstract}

\section{Introduction}

DFO addresses complex optimization problems where only noisy function evaluations are available through an oracle or black-box interface. This optimization framework has broad applications across diverse fields, including engineering design, black-box system optimization, and hyperparameter tuning in machine learning. For example, Google has developed an internal service, {\it Google Vizier}, based on DFO methods to optimize many machine learning models and complex systems \citep{Golovin2017GoogleVizier}. This service has executed millions of optimizations, accelerating numerous research and production systems at Google \citep{song2024viziergaussianprocessbandit}. For a comprehensive overview of DFO, refer to \cite{LarsonWild2019DFO}. In this paper, we focus on DFO via FD approximation, an important topic in simulation optimization \citep[see, e.g.,][]{ChauFu2014,Fu2015Gradient,hu2024convergence}, which has recently attracted substantial attention in machine learning \citep[see, e.g.,][]{shi2023numerical,Xuan2023thesis}.

FD approximation is a classical gradient estimation approach when the closed-form derivatives of functions are unavailable. This method involves generating samples at perturbed inputs. \cite{Fox1989Replication} and \cite{Zazanis1993Convergence} examine a batch-based FD estimator that uses a batch of independent and identically distributed (i.i.d.) samples generated at perturbed inputs. Specifically, they provide the theoretical optimal perturbation size given the batch size. However, unknown constants in the optimal perturbation size present a challenge for the practical use of the batch-based FD estimator. Thus, the estimator must either assume known constants or tune them as hyperparameters. To overcome this issue, \cite{Li2020Optimally} provide a two-stage procedure called Estimation-Minimization Central FD (EM-CFD). In the first stage, the optimal perturbation is estimated by some pilot samples; in the second stage, the FD estimator is constructed by additional samples generated at the estimated perturbation. Recently, \cite{Liang2024efficient} propose a novel FD method called correlation-induced FD (Cor-FD) to address the limitations of conventional FD methods. Unlike conventional FD methods that assume known optimal perturbation parameters, Cor-FD employs a sample-driven framework, dynamically estimating the optimal perturbation by combining bootstrap and regression techniques. The method's distinctive feature lies in its ability to leverage all available simulation samples by transforming them according to the estimated optimal perturbation, thereby generating correlated samples that enhance estimation accuracy. Also, \cite{Liang2024efficient} demonstrate that their proposed Cor-FD estimator possesses a reduced variance, and in some cases a reduced bias, compared to the conventional optimal FD estimator.

In simulation optimization, DFO via FD can track back to \cite{Kiefer1952Stochastic}, where they originally design a one-dimensional optimization problem and propose a gradient decent method that replaces the gradient with an FD estimator using only two function evaluations per iteration. This is called the KW algorithm. In the high-dimensional setting, \cite{Spall1992Multivariate} propose a more efficient alternative, known as the SPSA algorithm. This algorithm maintains the two-function-evaluation requirement regardless of dimensionality by employing a random perturbation vector, making it particularly suitable for large-scale optimization problems. To ensure algorithm convergence, both the step size and perturbation size must approach zero as iterations progress to infinity, and they must satisfy certain relationships. Notably, the performance of the KW and SPSA algorithms is sensitive to both step size and perturbation size, which must therefore be selected carefully.

Another line of research in DFO via FD sets a fixed perturbation size, with no requirement for the step size to tend to zero. As a result, the algorithm’s output converges to a neighborhood around the optimal solutions \citep{berahas2019derivative,shi2023numerical, bollapragada2024derivative}. Specifically, \cite{shi2023numerical} strongly recommends solving DFO problems via FD due to its natural compatibility with parallel computing. Moreover, it can be built based on the existing software, thereby avoiding the need to redesign the current DFO algorithms to handle general problems.

Most algorithms about DFO via FD apply the FD estimator with only two function evaluations per iteration to conserve samples during multiple iterations. Although this approach seems computationally efficient, it encounters several fundamental limitations. The gradient estimates obtained from such minimal sampling often suffer from significant inaccuracies, arising from the choice of perturbation parameters and inherent simulation noise. To ensure convergence under such imprecise gradient estimates, these methods must resort to diminishing step sizes, which often result in slow convergence rates and reduced practical efficiency. The batch-based FD estimator seems sample-consuming, yet it can approximate the gradient more accurately, providing a more accurate direction of gradient descent. This enhanced precision in gradient estimation enables the use of constant step sizes throughout the optimization process, eliminating the necessity of diminishing step lengths that are typically required by less accurate estimators.

In this paper, we apply the correlation-induced central FD (Cor-CFD) method to solve DFO problems and provide a numerical comparison with the aforementioned two classical algorithms, KW and SPSA, under various experimental settings, including low-dimensional and high-dimensional test problems with varying levels of noise and different function landscapes. Furthermore, we propose a hyperparameter optimization with real data. Specifically, we formulate the problem as a DFO problem, focusing on tuning the penalty parameter in ridge regression with real housing price data.
In low-dimensional problems, the numerical results show that Cor-CFD maintains stable convergence behavior, while KW often exhibits oscillation‌ in its optimization trajectory. The Cor-CFD method generally outperforms the KW in terms of solution quality. 
When tackling high-dimensional problems, Cor-CFD demonstrates substantially faster convergence rates compared to SPSA.
For the hyperparameter optimization problem, Cor-CFD converges more rapidly than KW.

The rest of this paper is organized as follows. 
Section \ref{section2} lays down the gradient-based optimization framework, introducing two mainstream approaches for gradient estimation in stochastic environments: methods using minimal samples per iteration (KW and SPSA) and method employing larger batch sizes for more accurate gradient estimation (Cor-CFD).
Section \ref{section3} validates the proposed method through numerical experiments, comparing it to the KW algorithm in one-dimensional cases and hyperparameter optimization, as well as the SPSA algorithm in high-dimensional case. The results highlight the batch-based method's superior performance in both convergence stability and computational efficiency, with Section \ref{section4} summarizing findings and future directions.

\section{Stochastic Optimization}\label{section2}

The stochastic optimization problem of interest is
\begin{align}\label{eq:SO_Problem}
    \min_{\bx\in \cX} \mu (\bx) = \E [Y(\bx)],
\end{align}
where $\bx=(x_1, \dots, x_d)\in \cX \subset \R^d$ denotes the decision variable, $\mu(\bx)$ denotes the mean response, and $Y(\bx)$ represents the realization of $\mu(\bx)$. Specifically, $Y(\bx)$ is expressed as $Y(\bx) = \mu(\bx) + \epsilon(\bx)$, where $\epsilon(\bx)$ is the random error.

To solve the optimization problem \eqref{eq:SO_Problem}, a well-known approach is the so-called stochastic approximation \citep{Robbins1951Stochastic} or stochastic gradient descent in machine learning if the unbiased gradient estimator exists, i.e., $\E[\nabla Y(\bx)] = \nabla \mu(\bx)$. The decision variable is updated iteratively based on the gradient of the objective function: 
\begin{align}\label{eq:GD}
    \bx_{k+1} = \bx_k - a_k \nabla Y(\bx_k), 
\end{align} 
where $a_k > 0$ is the step size, and $\nabla Y(\bx_k)$ represents the gradient estimator at iteration $k$. 

In many real-world problems, however, the unbiased gradient estimator $\nabla{Y}(\bx_k)$ is unavailable due to the complexity of systems. Furthermore, obtaining unbiased gradient information in practice, especially in black-box optimization or DFO, is time-consuming or even impossible. For example, \citep{Golovin2017GoogleVizier} state that ``Any sufficiently complex system acts as a black box when it becomes easier to experiment with than to understand." Therefore, it is important to give the optimization algorithm only with the simulation oracle.  Recently, \cite{shi2023numerical} argue that the FD gradient approximation, often overlooked, should be recommended in the DFO literature.

Specifically, \cite{shi2023numerical} suggest replacing $\nabla{Y}(\bx_k)$ in \eqref{eq:GD} with the FD estimator. Notably, they apply a second-order algorithm, substituting the decent direction $\nabla{Y}(\bx_k)$ in \eqref{eq:GD} with $H_k \nabla{Y}(\bx_k)$, where $H_k$ represents the inverse of the Hessian matrix at $k$-th iteration, effectively combining the FD method with the quasi-Newton method. For simplicity in this paper, $H_k$ is retained as an identity matrix, and only first-order optimization algorithms are considered.
Although the operation is simple to implement, the algorithm only converges to a region near the optimal solution \citep{berahas2019derivative}. As the noise increases, the region will expand, ultimately causing the optimization algorithm to deteriorate. Two main strategies can ensure algorithm convergence. The first involves using a diminishing step size ($a_k \to 0$) to reduce the impact of variance from the FD estimator, as applied in KW and SPSA. The second approach increases the number of samples (used for gradient estimation) to enhance the accuracy of the FD estimator. Next, we will provide elaborate explanations of these methods and compare them through experiments.

\subsection{Kiefer-Wolfowitz (KW) Algorithm}

The KW algorithm, originally developed for one-dimensional problems, is one of the earliest stochastic optimization algorithms based on finite differences. It estimates the gradient at each iteration using a CFD approach. Specifically, for a one-dimensional function, the gradient is approximated by:
\begin{align*} 
g_{k} = \frac{Y(x_k + c_k) - Y(x_k - c_k)}{2c_k}, 
\end{align*} 
where $x_k\in \R$ and $c_k$ is a perturbation parameter. The KW algorithm is 
\begin{align*}
    x_{k+1} = x_k - a_k g_k, 
\end{align*}
where $a_k>0$ is the step size.

To ensure the convergence of $x_k$ to the optimal value, the step size $a_k$ and the perturbation $c_k$ have to converge to 0 as $k\to\infty$ and satisfy some certain relationships. \cite{Kiefer1952Stochastic} provide the optimal convergence rates of $a_k$ and $c_k$. However, the constants in the rates are typically unknown. Experimental studies show that the performance of the KW algorithm is highly sensitive to the choices of $a_k$ and $c_k$. To overcome this issue, \cite{broadie2011general} propose a Scaled-and-Shifted KW (SSKW) for adaptively adjusting $c_k$ and $a_k$ to appropriate values by introducing extra 9 hyperparameters \citep[see, e.g.,][]{ChauFu2014}.

\subsection{Simultaneous Perturbation Stochastic Approximation (SPSA) Algorithm}

The SPSA algorithm is developed as an extension of KW for multi-dimensional optimization. SPSA introduces a key innovation by using simultaneous perturbation to estimate the gradient: rather than perturbing each variable individually, it applies random perturbations to all variables at once. This approach allows SPSA to approximate the gradient with only two function evaluations, regardless of the number of dimensions.

The gradient in SPSA is approximated using: 
\begin{align} 
g_k = \frac{Y(\bx_k + c_k \Delta_k) - Y(\bx_k - c_k \Delta_k)}{2 c_k} \Delta_k^{-1}, 
\end{align} 
where $c_k$ is a perturbation parameter, $\Delta_k$ is a random perturbation vector, and $\Delta_k^{-1}$ is an element-wise reciprocal of each component in the vector $\Delta_k$. 
The perturbation vector $\Delta_k$ is typically composed of independent random variables drawn from a symmetric Bernoulli distribution. Specifically, for each dimension $i$, the component $\Delta_{k,i}$ of the vector $\Delta_k$ is defined as $\pm 1$, meaning that each $\Delta_{k,i}$ independently takes a value of $+1$ or $-1$ with equal probability.
The SPSA algorithm is 
\begin{align*}
    \bx_{k+1} = \bx_k - a_k g_k, 
\end{align*}
where $a_k>0$ is the step size.

SPSA’s simultaneous perturbation strategy significantly reduces the number of function evaluations consumed per iteration, making it particularly efficient for high-dimensional optimization problems. However, despite this computational advantage, SPSA may experience slow convergence due to high variance in its gradient estimates, especially in noisy environments. This variance can impede convergence, making SPSA slower in practice compared to methods that yield more precise gradient estimates. Furthermore, due to the inherent inaccuracy in gradient estimation, the SPSA algorithm requires the step size $a_k$ and the perturbation parameter $c_k$ to approach zero for convergence. The choice of step size $a_k$ and perturbation size $c_k$ significantly influences SPSA's performance, demanding careful tuning to achieve optimal convergence rates.

\subsection{Batch-Based Algorithm}

In this section, we introduce an FD method proposed by \cite{Liang2024efficient}, called the Cor-FD method, and apply it into \eqref{eq:GD}. The Cor-FD method employs a batch of samples for gradient estimation.

For simplicity, we consider the $k$-th iteration with $d = 1$. When $d > 1$, the same operation can be applied to each coordinate direction. Furthermore, assume that $\mu(x)$ is thrice continuously differentiable in a neighborhood of $x_k$, where $x_k$ is the current solution, and the noise at $x_k$ is non-zero, i.e., $\Var[\epsilon(x_k)] > 0$. 
Using the CFD method with $n_k$ sample pairs to estimate $\mu'(x_k)$, where $n_k$ denotes the batch size at $k$-th iteration and each sample pair involves two function evaluations, the gradient estimator with perturbation size $c$ can be expressed as:
\begin{align*}
    g_{n_k, c} = \frac{1}{n_k}\sum_{i=1}^{n_k}\frac{Y_i(x_k + c) - Y_i(x_k - c)}{2c}.
\end{align*}
It is straightforward to calculate that the expectation of $g_{n_k,c}$ is
\begin{align*}
    \E[g_{n_k,c}] = \frac{\mu(x_k + c) - \mu(x_k - c)}{2c} = \mu'(x_k) + B c^2 + o(c^2),
\end{align*}
where $B = \mu^{(3)}(x_k)/6$, $\mu^{(3)}(x_k)$ is the third derivative, and the second equality is due to the Taylor expansion of $\mu(x_k + c)$ and $\mu(x_k - c)$ at $x_k$. The variance of $g_{n_k,c}$ is given by
\begin{align*}
    \Var[g_{n_k, c}] = \frac{\Var[\epsilon(x_k + c)] + \Var[\epsilon(x_k - c)]}{4n_k c^2} = \frac{\Var[\epsilon(x_k)] + o(1)}{2n_k c^2}.
\end{align*}
Then the mean squared error (MSE, equal to $\mbox{Bias}^2 + \mbox{Variance}$) of $g_{n_k, c}$ is
\begin{align*}
    \MSE[g_{n_k, c}] = ( B + o(1) )^2 c^4 + \frac{\Var[\epsilon(x_k)] + o(1)}{2n_k c^2}.
\end{align*}
The optimal perturbation to minimize the MSE of the estimator is $c_k = (\Var[\epsilon(x_k)] / 4 n_k B^2)^{1/6}$ \citep{Fox1989Replication, Zazanis1993Convergence}. Note that this constant depends on the unknown constants $\Var[\epsilon(x_k)]$ and $B$, which could be more difficult to estimate than $\mu'(x_k)$.

To choose an appropriate perturbation and then estimate the gradient $\mu'(x_k)$ efficiently, the Cor-CFD method samples $R$ different pilot perturbations $c_{k,1}, c_{k,2}, ..., c_{k,R}$ from a distribution $\mathcal{P}_0$ with variance $O\left(n_{k}^{-1/5}\right)$. For each $c_{k,r} (r = 1, 2, ..., R)$, $b_k$ sample pairs $\left(Y_i(x_k + c_{k,r}), Y_i(x_k - c_{k,r})\right) (i = 1, 2, ..., b_k)$ are generated, assuming without loss of generality that $R$ is divisible by $n_k$ and $n_k = b_k R$. This yields $R$ distinct CFD estimators:
\begin{align*}
    g_{b_k,c_{k,r}} = \frac{1}{b_k}\sum_{i=1}^{b_k}\frac{Y_i(x_k + c_{k,r}) - Y_i(x_k - c_{k,r})}{2c_{k,r}}, \quad r = 1, 2, ..., R.
\end{align*}

By employing the bootstrap, we can estimate the expectation $\E[g_{b_k,c_{k,r}}]$ and variance $\Var[g_{b_k, c_{k,r}}]$ for each $r$, denoted by $\E_*[g_{b_k,c_{k,r}}]$ and $\Var_*[g_{b_k, c_{k,r}}]$, respectively. 
Regressing $\left[\E_*[g_{b_k, c_{k,1}}],...,\E_*[g_{b_k, c_{k,R}}]\right]$ on $[1,...,1]$ and $\left[ c_{k,1}^2,...,c_{k,R}^2 \right]$ gives estimates $\left[\widehat{\mu}'(x_k), \widehat{B} \right]$, and regressing $\left[\Var_*[g_{b_k, c_{k,1}}],..., \Var_*[g_{b_k, c_{k,R}}]\right]$ on $\left[ 1/2b_k c_{k,1}^2,...,1/2b_k c_{k,R}^2 \right]$ gives estimate $\widehat{\Var}[\epsilon(x_k)]$. Then, the best perturbation for $n_k$ is 
\begin{align*}
    \hat{c}_k = \left( \frac{\widehat{\Var}[\epsilon(x_k)]}{4n_k^2 \widehat{B}^2} \right)^{1/6}.
\end{align*}

To conserve samples, we recycle them by adjusting their location and scale based on the expectation and standard deviation of $\left(Y(x_k + \hat{c}_k) - Y(x_k - \hat{c}_k)\right)/(2\hat{c}_k)$. For any $r = 1,..., R$ and $i=1,..., b_k$, we transform $(Y_i(x_k + c_{k,r}) - Y_i(x_k - c_{k,r}))/2c_{k,r}$ to
\begin{align}\label{eq:individual_CorCFD}
    \frac{c_{k,r}}{\hat{c}_k} \left[\frac{Y_i(x_k + c_{k,r}) - Y_i(x_k - c_{k,r})}{2c_{k,r}} - \widehat{\mu}'(x_k) - \widehat{B} c_{k,r}^2 \right] + \widehat{\mu}'(x_k) + \widehat{B} c_{k,r}^2.
\end{align}

Finally, the Cor-CFD estimator is formulated as:
\begin{align*}
    g_{n_k}^{cor}(x_k) =  \frac{1}{n_k} \sum_{r=1}^{R} \sum_{i=1}^{b_k} \eqref{eq:individual_CorCFD}. 
\end{align*}

A more detailed procedure can be found in \cite{Liang2024efficient}. They present the following theoretical result, revealing that the Cor-CFD estimator achieves a reduction in variance and a possible decrease in bias compared to the optimal CFD estimator. Consequently, the Cor-CFD estimator consistently performs nearly as well as (or even outperforms) the optimal CFD estimator.
\begin{theorem}[Theorem 4 in \cite{Liang2024efficient}]
    Assume that $\mu(x)$ is fifth differentiable at $x_k$ with non-zero fifth derivative, and $\Var[\epsilon(x)] > 0$ is continuous at $x_k$. For any $r = 1,...,R$ $(R \geq 2)$, let $c_{k,r} = t_{k,r} n_k^{-1/10}$ $(c_{k,r} \neq 0)$ and for any $s \neq r$, $c_{k,s} \neq c_{k,r}$. If $n_k \to \infty$, then we have
	\begin{align*}
	&\E[g_{n_k}^{cor}(x_k)] = \mu'(x_k) + \left(\frac{B\Var[\epsilon(x_k)]}{4n}\right)^{\frac{1}{3}} + \left(\frac{4B^2}{\Var[\epsilon(x_k)]}\right)^{\frac{1}{6}}\frac{D \Lambda}{\sqrt{R}}n_k^{-\frac{1}{3}} + o\left(n_k^{-\frac{1}{3}}\right),\\
	&\Var[g_{n_k}^{cor}(x_k)] = \left(\frac{B^2 \Var[\epsilon(x_k)]^2}{2n_k^2}\right)^{\frac{1}{3}} + \left(\frac{B^2 \Var[\epsilon(x_k)]^2}{2}\right)^{\frac{1}{3}}\frac{q - R}{R}n_k^{-\frac{2}{3}} + o\left(n_k^{-\frac{2}{3}}\right),
\end{align*}
where $D = \mu^{(5)}(x_k)/120$, ${\boldsymbol{t}_k} = [|t_{k,1}|,...,|t_{k,R}|]^{\top}$, ${\boldsymbol{t}_k^4} = \left[t_{k,1}^4,...,t_{k,R}^4\right]^{\top}$, $\Lambda = \boldsymbol{t}_k^{\top}P\boldsymbol{t}_k^4$, ${\boldsymbol{P}}=\boldsymbol{I} - \boldsymbol{X}_e(\boldsymbol{X}_e^{\top}\boldsymbol{X}_e)^{-1}\boldsymbol{X}_e^{\top}$, $\boldsymbol{X}_e$ is a $R\times 2$ matrix with the first and second columns being $[1,...,1]^{\top}$ and $[t_{k,1},...,t_{k,R}]^{\top}$, respectively, $q = \left\|{\rm{Diag}}(\boldsymbol{t}_k^{-1})\boldsymbol{Pt}_k\right\|_2^2$ and ${\rm{Diag}}(\boldsymbol{t}_k^{-1}) = {\rm{Diag}}\left(1/|t_{k,1}|,...,1/|t_{k,R}| \right)$.
\end{theorem}

Next, we embed the Cor-CFD method into the gradient-based optimization algorithm for $d \geq 1$. Specifically, we substitute $\nabla Y(\bx_k)$ in \eqref{eq:GD} with the Cor-CFD estimator to update the decision variable. Notably, {\it while the Cor-CFD method can also be embedded within quasi-Newton methods, we omit this case here to ensure a fair comparison with first-order algorithms, including KW and SPSA}. At $k$-th iteration, two questions must be addressed: selecting the step size $a_k$ and determining the batch size $n_k$ for each coordinate. For the step size, we apply the stochastic Armijo condition \citep[see, e.g.,][]{berahas2019derivative, shi2023numerical},
\begin{align}\label{eq:Armijo}
    Y(\bx_{k} - a_{k} g_{k}) \leq Y(\bx_{k}) - l_1 a_k g_{k}^{\top} g_{k} + 2 \sqrt{\widehat{\Var}[\epsilon(\bx_{k})]}.
\end{align}
Specifically, we iteratively reduce $a_k \leftarrow l_2 a_{k}$ until the above condition holds. 
In the standard version of the stochastic Armijo condition, the term $\widehat{\Var}[\epsilon(\bx_{k})]$ is replaced with an upper bound on the variance. However, in Cor-CFD, the estimation of $\widehat{\Var}[\epsilon(\bx_{k})]$ can adaptively capture heteroskedasticity, providing additional flexibility and precision.
For the batch size, we allow $n_k$ to increase gradually as $k$ grows. Here, we set $n_k = \lfloor (n_0 + k)/R \rfloor \times R$, where $\lfloor \cdot \rfloor$ denotes the floor operation, rounding down to the nearest integer, and $R$ is divisible by $n_0$. Note that the setting is not optimal because $n_k$ depends on the true gradient $\nabla{\mu}(\bx_k)$ and noise level $\Var[\epsilon(\bx_{k})]$. For example, when $k$ is small, $\bx_k$ is far from the optimal point, making the optimal $n_k$ relatively small. Conversely, as $k$ grows large, the true gradient approaches 0 and the noise becomes prominent, thus increasing the optimal $n_k$. Because our objective here is to validate the effectiveness of the batch method, the precise determination of $n_k$ is left for future work. In this paper, we consider the Cor-CFD-gradient descent (Cor-CFD-GD) algorithm. The complete procedure is summarized in Algorithm \ref{alg:algorithm_SO}.

\begin{algorithm2e}[t!]
	\caption{Cor-CFD-GD Algorithm.}
	\label{alg:algorithm_SO}
	\BlankLine
	\textit{\textbf{Input: }} The number of total sample pairs $n$, the number of perturbation parameters $R$, the number of initial batch size $n_0$, initial perturbation generator $\mathcal{P}_0$, the line search parameters $(l_1,l_2)$, starting point $\bx_0$ and initial step length for line search $a_0 > 0$.
	
	\textit{\textbf{Initialization: }} Set the iteration counter $k = 0$ and the function evaluation counter $n_{count} = 0$.
	
	\textit{\textbf{While $n_{count} < 2n$ do}} 
	\begin{enumerate}
            \item Set $n_{k} = \lfloor(n_0 + k)/R \rfloor \times R$ and obtain the estimator $g_k$ using the Cor-CFD algorithm.
            \item Set $n_{count} = n_{count} + 2dn_k$.
		\item Use stochastic Armijo line search \eqref{eq:Armijo} to find an appropriate step length $0 < a_k \leq a_0$ and set $n_{count} = n_{count} + n_{ls}$, where $n_{ls}$ denotes the function evaluation counter during the line search procedure.
		\item Update $\bx_{k+1} = \bx_k - a_k g_k$.
		\item Set $k = k + 1$.
	\end{enumerate}
	
	\textit{\textbf{Output: }} The ultimate estimate $\bx_{k+1}$.
\end{algorithm2e}

\section{Experimental Studies}\label{section3}

\subsection{One Dimensional Examples with KW}\label{sec:KW}

In this section, we present two examples from the numerical experiments in \cite{broadie2011general}, designed to evaluate algorithm performance under different levels of artificial noise. 
The model $Y(x) = \mu(x) + \epsilon(x)$ is considered, where $x$ is a scalar, and the noise term $\epsilon(x)$ is a zero-mean Gaussian noise with variance $\sigma^2$. 
\begin{itemize}
    \item $\mu(x) = x^4$, with noise levels $\sigma \in \{0.1, 1, 10\}$.
    \item $\mu(x) = -100\cos(\pi x / 100)$, with noise levels $\sigma \in \{ 1, 10, 100 \}$.
\end{itemize}
For both examples, the domain is set to be $\mathcal{X} = [-50, 50]$ with a starting point of $x_0 = 30$. 
These two examples are both convex. Although it is well-known that the KW algorithm may work better on strongly convex functions, the numerical results support the potential advantages of Cor-CFD, compared with KWSA and SPSA algorithms. 

To assess the impact of batch on stochastic approximation, we compare the KW algorithm with two batch-based stochastic approximation methods: Cor-CFD-GD and EM-CFD-GD. The Cor-CFD-GD is presented in Algorithm \ref{alg:algorithm_SO}. The EM-CFD-GD is an optimization algorithm that follows the structure of Algorithm \ref{alg:algorithm_SO}, using the EM-CFD proposed by \cite{Li2020Optimally} to estimate the gradient. To ensure convergence of the gradient estimator, the batch size is set as $n_k = n_0 + 2 \times k$.
To maintain solutions within the interval $\mathcal{X}$, we apply a truncation technique at each iteration. Specifically, at the $k$-th iteration, the update step is defined as $x_{k+1} = \Pi_{\mathcal{X}}\left(x_k - a_k g_k\right)$, where $\Pi_{\mathcal{X}}$ denotes the projection operator onto $\mathcal{X}$. The projection operator maps any point $x$ into the domain $\mathcal{X}$ by selecting the closest point within $\mathcal{X}$ to $x$.
\begin{itemize}
    \item For the KW algorithm, we use tuning sequences $a_k = a/k $, $ c_k = c/k^{1/4}$, setting $a = c = 1$ as recommended in many studies \citep[see, e.g.,][]{broadie2011general}. 
    \item For the Cor-CFD-GD algorithm, the initial perturbation generator $\mathcal{P}_0$ is chosen as a truncated normal distribution with mean 0 and variance 1, truncated to the range $[0.1, \infty)$ to prevent excessively small perturbation. The initial batch size is set to $n_0=20$, and the number of perturbation parameters is $R=5$. For the Armijo condition, we set $(l_1, l_2) = (10^{-4}, 0.5)$, the initial step length $a_0 = 1$.
    \item For the EM-CFD-GD algorithm, the initial perturbation generator $\mathcal{P}_0$ is set to a normal distribution with mean 0 and variance 1, and the initial batch size is $n_0=20$. The sample ratio in stage 1 is set to be $0.5$. The Armijo condition remains the same as that of the Cor-CFD-GD algorithm.
\end{itemize}

We compare the root mean squared error (RMSE) of the solution gap and the oscillatory period length. 
\begin{itemize}
    \item RMSE(Solution Gap) is defined as the RMSE of the distance between the solution point and the optimal point $0$.
    \item The oscillatory period is defined as the number of sample pairs until the algorithm stops oscillating between boundary points. Specifically, in this section, this period is represented by the cardinality of the set $ \{k \geq 2 : (x_k = 50 \ \text{and} \ x_{k-1} = -50) \ \text{or} \ (x_k = -50 \ \text{and} \ x_{k-1} = 50)\}$.
\end{itemize}

Tables \ref{tab:example1} and \ref{tab:example2} present the results for $\mu(x) = x^4$ and $\mu(x) = -100\cos (\pi x / 100)$, respectively, each based on 200 replications. Both tables report the RMSE(Solution Gap) and the length of the oscillatory period for different methods at 100, 1000, and 10000 sample pairs under varying noise levels. We also report the 5th, 50th, and 95th percentiles of the length of the oscillatory period, denoted by ``5\%'', ``Median'', and ``95\%'', respectively. The oscillatory period length is recorded based on 10000 sample pairs. 
Table \ref{tab:example1} shows that for sample pairs of 100 and 1000, the KW algorithm oscillates near the boundaries, while the Cor-CFD-GD and EM-CFD-GD algorithms remain within the boundaries, with RMSE steadily decreasing as the number of sample pairs increases. This highlights the importance of accurate gradient estimation in gradient-based optimization algorithms. 
The batch-based methods employ line search to determine the step size, which prevent oscillatory behavior.
Moreover, when the sample pairs are 10000, the KW algorithm's performance shows little variation across different values of $\sigma$. This occurs because, once the KW algorithm stops crossing the boundaries and begins to converge, the small value of $c_k$ leads to significant FD errors that dominate $\sigma$. It is worth noting that the Cor-CFD-GD algorithm underperforms compared to the KW algorithm when $\sigma = 10$ and sample pairs are 10000. This may result from a slow increase in batch size, leading to insufficient precision in gradient estimation and thus a slower descent rate.
Furthermore, the EM-CFD-GD algorithm underperforms compared to the Cor-CFD-GD algorithm. This is because the EM-CFD-GD algorithm allocates half of the samples only for estimating the optimal perturbation, without reusing them for gradient estimation. This results in insufficient utilization of the available samples.

\begin{table}[t]
\renewcommand\arraystretch{1.25}
\centering
\caption{Comparison of the Cor-CFD-GD algorithm, the EM-CFD-GD algorithm and the KW algorithm for $\mu(x) = x^4$.}
\label{tab:example1}
\vspace{1em}
\begin{tabular}{cccccccc}
\toprule

 &  & \multicolumn{3}{c}{RMSE(Solution Gap)} & \multicolumn{3}{c}{Length of oscillatory period} \\
\cmidrule(lr){3-5} \cmidrule(lr){6-8}
$\sigma$ & Method & \multicolumn{1}{r}{100} & \multicolumn{1}{r}{1,000} & \multicolumn{1}{r}{10,000} & \multicolumn{1}{r}{5\%} & \multicolumn{1}{r}{Median} & \multicolumn{1}{r}{95\%} \\
\midrule
\multirow{3}{*}{0.1} & Cor-CFD-GD & \multicolumn{1}{r}{0.10} & \multicolumn{1}{r}{0.01} & \multicolumn{1}{r}{0.01} & \multicolumn{1}{r}{0} & \multicolumn{1}{r}{0} & \multicolumn{1}{r}{0} \\
& EM-CFD-GD       & \multicolumn{1}{r}{1.67}  & \multicolumn{1}{r}{0.03}  & \multicolumn{1}{r}{0.02} & \multicolumn{1}{r}{0} & \multicolumn{1}{r}{0} & \multicolumn{1}{r}{0} \\
& KW       & \multicolumn{1}{r}{50.00}  & \multicolumn{1}{r}{50.00}  & \multicolumn{1}{r}{0.42} & \multicolumn{1}{r}{5000} & \multicolumn{1}{r}{5000} & \multicolumn{1}{r}{5000} \\
\multirow{3}{*}{1}   & Cor-CFD-GD & \multicolumn{1}{r}{0.23} & \multicolumn{1}{r}{0.11} & \multicolumn{1}{r}{0.06} & \multicolumn{1}{r}{0} & \multicolumn{1}{r}{0} & \multicolumn{1}{r}{0} \\
& EM-CFD-GD       & \multicolumn{1}{r}{1.80}  & \multicolumn{1}{r}{0.48}  & \multicolumn{1}{r}{0.19} & \multicolumn{1}{r}{0} & \multicolumn{1}{r}{0} & \multicolumn{1}{r}{0} \\
& KW       & \multicolumn{1}{r}{50.00}  & \multicolumn{1}{r}{50.00}  & \multicolumn{1}{r}{0.42} & \multicolumn{1}{r}{4999} & \multicolumn{1}{r}{5000} & \multicolumn{1}{r}{5000} \\
\multirow{3}{*}{10}  & Cor-CFD-GD & \multicolumn{1}{r}{1.21} & \multicolumn{1}{r}{1.21} & \multicolumn{1}{r}{1.12} & \multicolumn{1}{r}{0} & \multicolumn{1}{r}{0} & \multicolumn{1}{r}{0} \\
& EM-CFD-GD       & \multicolumn{1}{r}{1.61}  & \multicolumn{1}{r}{1.29}  & \multicolumn{1}{r}{1.23} & \multicolumn{1}{r}{0} & \multicolumn{1}{r}{0} & \multicolumn{1}{r}{0} \\
& KW       & \multicolumn{1}{r}{50.00}  & \multicolumn{1}{r}{50.00}  & \multicolumn{1}{r}{0.43} & \multicolumn{1}{r}{4999} & \multicolumn{1}{r}{5000} & \multicolumn{1}{r}{5000} \\

\bottomrule
\end{tabular}
\end{table}

As shown in Table \ref{tab:example2}, the Cor-CFD-GD algorithm consistently outperforms the KW algorithm and the EM-CFD-GD algorithm, achieving faster descent. For the case where $\sigma=100$, the gradient's magnitude becomes small relative to noise variance, making optimization more challenging. In this setting, the KW algorithm converges slowly, while the Cor-CFD-GD algorithm rarely crosses the boundary and continues its descent. The slower convergence of the EM-CFD-GD algorithm can be attributed to its insufficient utilization of samples. This highlights the advantage of using a batch of samples for gradient estimation in the gradient-based optimization algorithm. Although it reduces the iteration steps, each iteration progresses rapidly toward the optimum, showing great promise in high-noise settings.

\begin{table}[t]
\renewcommand\arraystretch{1.25}
\centering
\caption{Comparison of the Cor-CFD-GD algorithm, the EM-CFD-GD algorithm and the KW algorithm for $\mu(x) = -100\cos (\pi x / 100)$.}
\label{tab:example2}
\vspace{1em}
\begin{tabular}{cccccccc}
\toprule
 &  & \multicolumn{3}{c}{RMSE(Solution Gap)} & \multicolumn{3}{c}{Length of oscillatory period} \\
\cmidrule(lr){3-5} \cmidrule(lr){6-8}
$\sigma$ & Method & \multicolumn{1}{r}{100} & \multicolumn{1}{r}{1,000} & \multicolumn{1}{r}{10,000} & \multicolumn{1}{r}{5\%} & \multicolumn{1}{r}{Median} & \multicolumn{1}{r}{95\%} \\
\midrule
\multirow{3}{*}{1}   & Cor-CFD-GD & \multicolumn{1}{r}{20.79} & \multicolumn{1}{r}{1.56} & \multicolumn{1}{r}{0.10} & \multicolumn{1}{r}{0} & \multicolumn{1}{r}{0} & \multicolumn{1}{r}{0} \\
& EM-CFD-GD       & \multicolumn{1}{r}{20.80} & \multicolumn{1}{r}{3.12} & \multicolumn{1}{r}{0.20} & \multicolumn{1}{r}{0} & \multicolumn{1}{r}{0} & \multicolumn{1}{r}{0} \\
& KW       & \multicolumn{1}{r}{18.73} & \multicolumn{1}{r}{15.05} & \multicolumn{1}{r}{12.10} & \multicolumn{1}{r}{0} & \multicolumn{1}{r}{0} & \multicolumn{1}{r}{0} \\
\multirow{3}{*}{10}  & Cor-CFD-GD & \multicolumn{1}{r}{20.92} & \multicolumn{1}{r}{2.43} & \multicolumn{1}{r}{1.11} & \multicolumn{1}{r}{0} & \multicolumn{1}{r}{0} & \multicolumn{1}{r}{0} \\
& EM-CFD-GD       & \multicolumn{1}{r}{21.33} & \multicolumn{1}{r}{5.24} & \multicolumn{1}{r}{2.16} & \multicolumn{1}{r}{0} & \multicolumn{1}{r}{0} & \multicolumn{1}{r}{0} \\
& KW       & \multicolumn{1}{r}{21.31} & \multicolumn{1}{r}{17.70} & \multicolumn{1}{r}{14.06} & \multicolumn{1}{r}{0} & \multicolumn{1}{r}{0} & \multicolumn{1}{r}{0} \\
\multirow{3}{*}{100} & Cor-CFD-GD & \multicolumn{1}{r}{25.78} & \multicolumn{1}{r}{16.88} & \multicolumn{1}{r}{10.75} & \multicolumn{1}{r}{0} & \multicolumn{1}{r}{0} & \multicolumn{1}{r}{0} \\
& EM-CFD-GD       & \multicolumn{1}{r}{34.48} & \multicolumn{1}{r}{27.03} & \multicolumn{1}{r}{21.82} & \multicolumn{1}{r}{0} & \multicolumn{1}{r}{0} & \multicolumn{1}{r}{0} \\
& KW       & \multicolumn{1}{r}{29.39} & \multicolumn{1}{r}{26.40} & \multicolumn{1}{r}{24.43} & \multicolumn{1}{r}{1} & \multicolumn{1}{r}{9} & \multicolumn{1}{r}{117} \\
\bottomrule
\end{tabular}
\end{table}

\begin{figure}[t!]
      \includegraphics[width=1\linewidth]{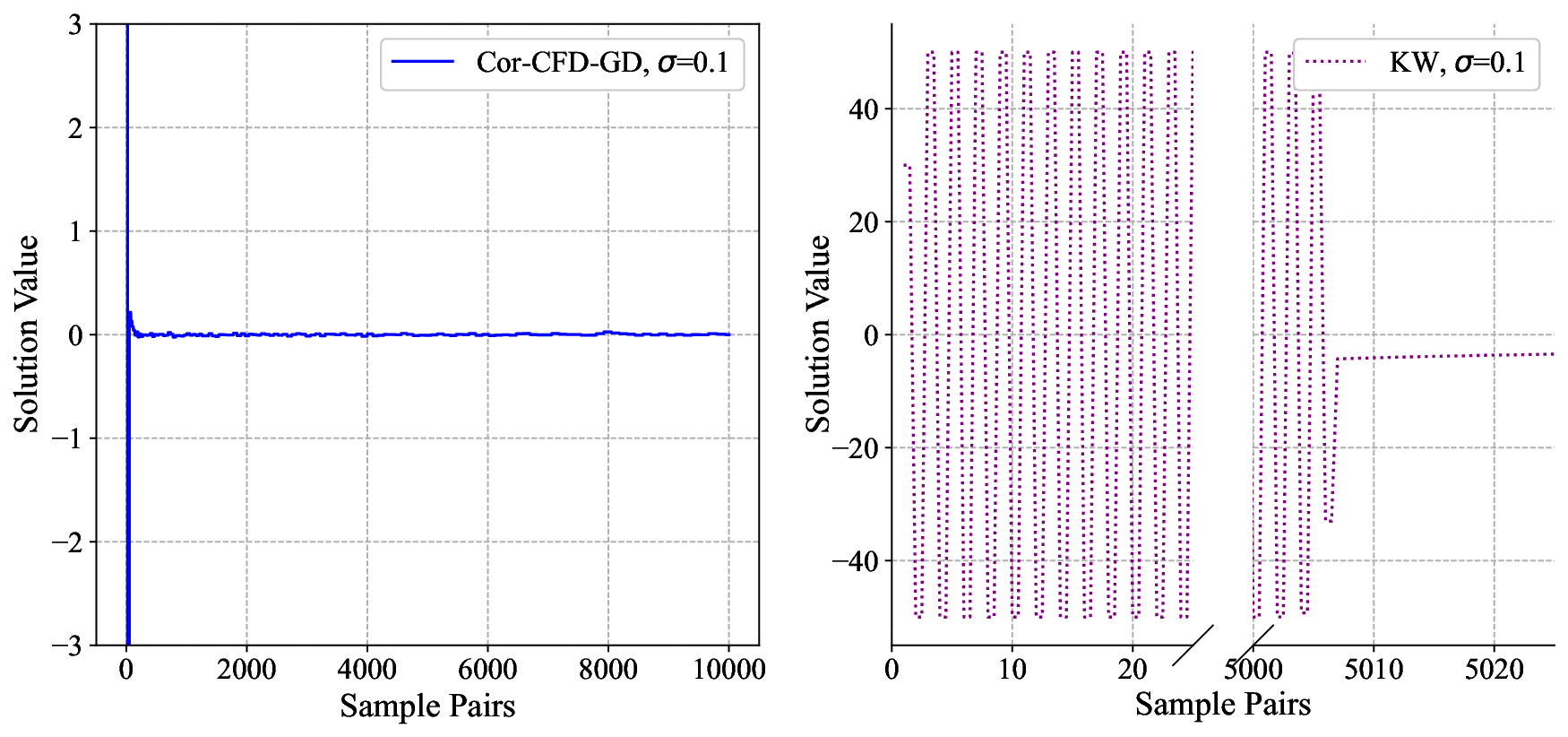}
      \caption{Solution value comparison between Cor-CFD-GD and KW algorithms across sample pairs for $\mu(x) = x^4$ with noise level $\sigma=0.1$.}
      \hspace*{-0.2cm}
      \label{fig:func_1_01}
\end{figure}

\begin{figure}[t!]
      \includegraphics[width=1\linewidth]{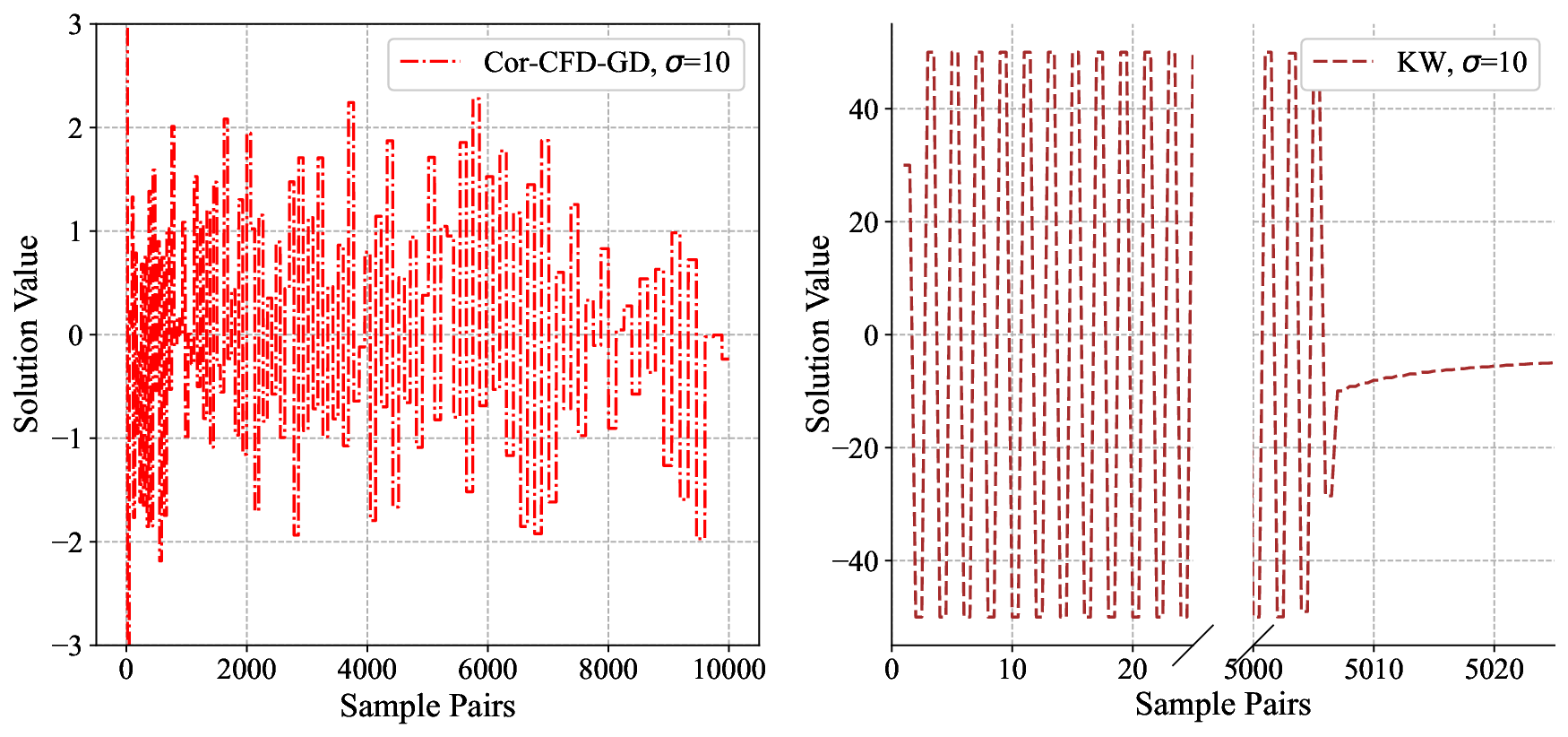}
      \caption{Solution value comparison between Cor-CFD-GD and KW algorithms across sample pairs for $\mu(x) = x^4$ with noise level $\sigma = 10$.}
      \hspace*{-0.2cm}
      \label{fig:func_1_10}
\end{figure}

\begin{figure}[t!]
      \includegraphics[width=1\linewidth]{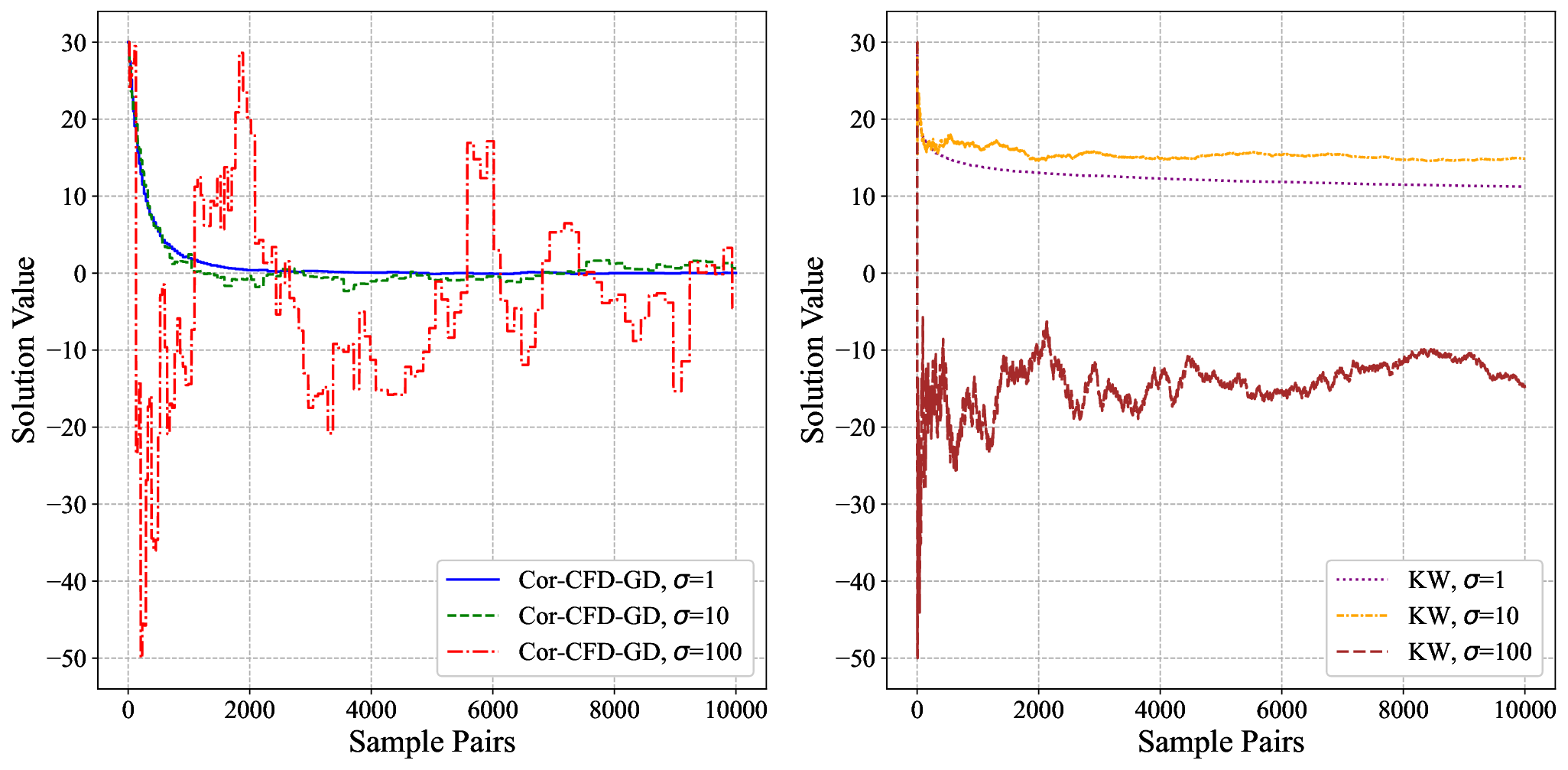}
            \caption{Solution value comparison between Cor-CFD-GD and KW algorithms across sample pairs for $\mu(x) = -100\cos (\pi x / 100)$.}
      \label{fig:func_2}
\end{figure}

To clarify the optimization results over time, Figs. \ref{fig:func_1_01} and \ref{fig:func_1_10} show the iterative processes of the Cor-CFD-GD and the KW algorithms for $\mu(x) = x^4$, while Fig. \ref{fig:func_2} shows for $\mu(x) = -100\cos(\pi x/100)$. The iterative process for $\sigma = 1$ is similar to that for $\sigma=0.1$ for $\mu(x) = x^4$, so it is not shown here.
From Figs. \ref{fig:func_1_01} and \ref{fig:func_1_10}, when the step size is too large relative to the gradient, the KW algorithm exhibits oscillations between boundary points until the step size decreases to an appropriate level. Effective optimization begins only after 5000 sample pairs, when the step size is sufficiently reduced. Therefore, under a limited sample pair budget, the KW algorithm fails if the initial parameters are not suitable. In contrast, the Cor-CFD-GD algorithm remains effective with small samples, suggesting that allocating samples to enhance gradient estimation accuracy is beneficial. Additionally, we observe continuous fluctuations in the Cor-CFD-GD algorithm as noise levels rise. This fluctuation does not imply a lack of convergence but rather occurs because batch size increases too slowly. As noise levels increase, the batch size must be raised to sustain confidence in gradient estimation.

Similar results can be observed in Fig. \ref{fig:func_2}, where the Cor-CFD-GD algorithm oscillates around the optimal point $x=0$ when $\sigma = 100$. However, the magnitude of fluctuation gradually decreases, with a maximum fluctuation of 49.4 near 2500 sample pairs and 21.8 near 7500 sample pairs. Although the KW algorithm exhibits minimal fluctuation, it converges very slowly and does not achieve its theoretically optimal convergence rate. This observation aligns with the discussion in \cite{broadie2011general}. For instance, when $\sigma = 1$ and sample pairs range from 1000 to 10000, the RMSE converges at a rate of only $1/10$ order.

\subsection{A Multi-Dimensional Example with SPSA}

In this section, we consider the multi-dimension version of function 213 from \cite{schittkowski_more_1987}:
\begin{equation}\label{function213}
    \mu(\bx) = \sum_{i=1}^{d/2} \left[10(x_{2i} - x_{2i-1})^2 + (1 - x_{2i-1})^2 \right]^4,
\end{equation}
where $d$ is a positive even integer representing the dimension of the input vector $\bx$.

The objective function reaches its global minimum of $0$ at $\bx^* = (1, \dots, 1)$. The response surface is modeled as $Y(\bx) = \mu(\bx) + \epsilon(\bx)$, where $\epsilon(\bx)$ represents a zero-mean Gaussian noise with variance $\sigma^2$. Experiments are conducted with $d=64$ and varying noise levels $\sigma \in \{0.1, 1, 10\}$.
Following the recommendation of \cite{schittkowski_more_1987}, we initialize the algorithm at $\bx_0 = (3, 1, \dots, 3, 1)$, where the function value is approximately $10^8$ at this starting point.

\begin{figure}[h!]
      \includegraphics[trim={0 1cm 0 3cm}, clip, width=\textwidth]{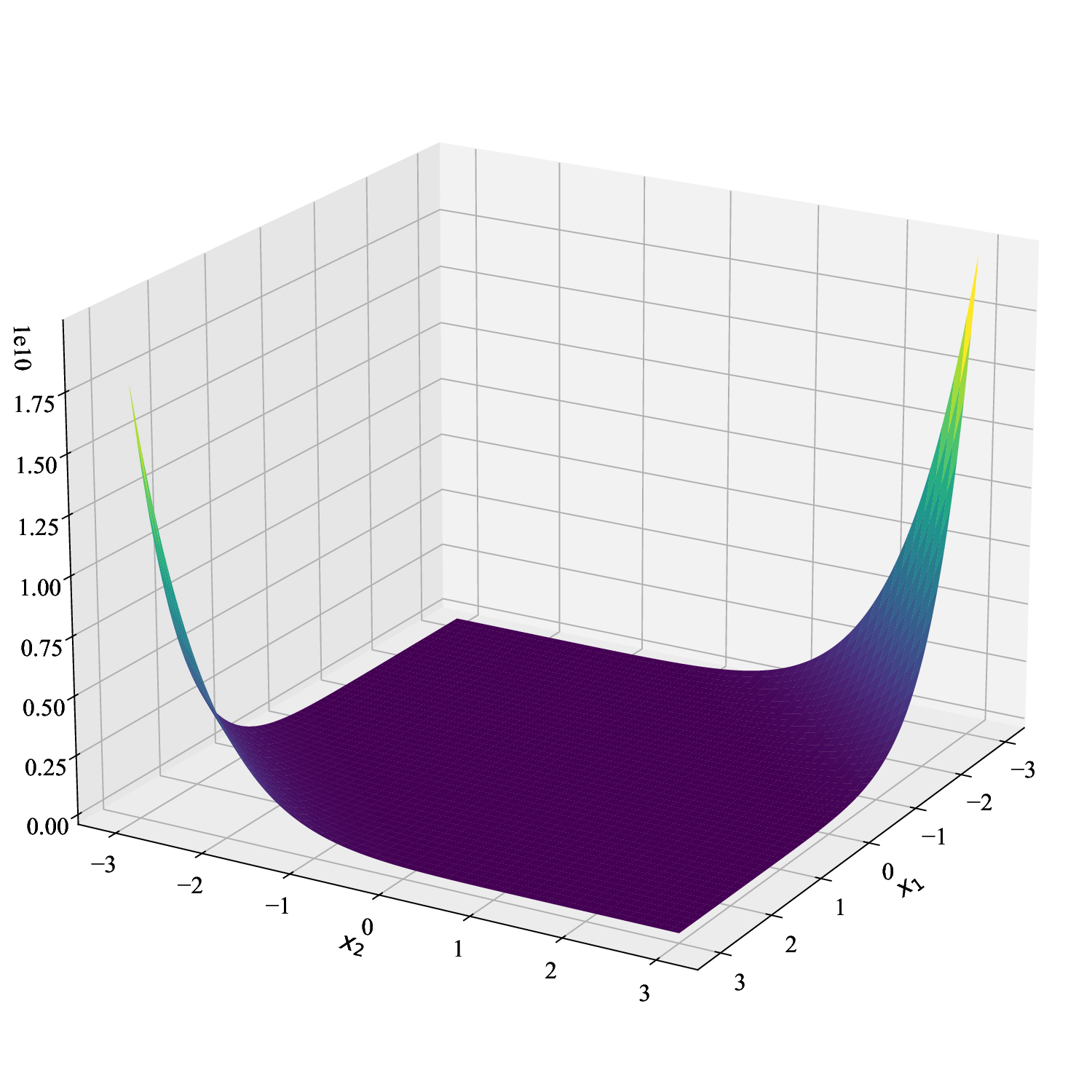}
      \caption{Image of function \eqref{function213} at $d=2$.}
      \label{fig:func213}
\end{figure}

Fig. \ref{fig:func213} shows the graph of function \eqref{function213} with $d=2$, illustrating its optimization challenges.
The graph exhibits a narrow, curved valley bordered by steep walls with large gradients. Near the optimum, the function becomes flat. Further from the optimum, the gradient magnitude increases, responding sensitively to small changes in $\bx$. This structure, characterized by steep outer regions and a relatively flat area near the minimum, poses substantial challenges for optimization.

As recommended in \cite{spall1998parameter}, the SPSA algorithm uses the parameters $a_k = a/(A + k + 1)^{0.602}$ and $c_k = c/(k+1)^{0.101}$, where $A$ is 10\% of the maximum number of sample pairs. To determine the optimal parameter configuration for SPSA under each experimental setting, we conduct a grid search over the parameter space. We evaluate combinations of the step size parameter $a \in \{ 10^{-1}, 10^{-2}, 10^{-3}, 10^{-4}, 10^{-5}, 10^{-6}, 10^{-7}, 10^{-8}, 10^{-9} \}$ and the perturbation parameter $c \in \{0.01, 0.1, 1, 2, 4\}$. 

The optimal parameter settings are as follows: for budgets of 1,000 and 5,000, the parameter $a$ is consistently set to $10^{-9}$ across all variance levels, while $c$ remains constant at 2. For the budget of 10,000, $a$ increases slightly to $10^{-8}$, yet $c$ continues to hold steady at 2 regardless of variance. This parameter stability can be attributed to the relatively large function values during the initial stages of optimization, which effectively overshadow the impact of variance. The noise introduced by varying levels of variance does not significantly affect the optimization trajectory, allowing these parameter settings to perform optimally across different settings.
Notably, increasing the step size $a$ causes the points to jump too far, preventing convergence, while decreasing $a$ results in very slow convergence.

For the Cor-CFD-GD algorithm, the initial perturbation generator $\mathcal{P}_0$ is set to be a truncated normal distribution with mean 0 and variance $0.1^2$, restricted to the range $[0.01, \infty)$, while all other parameters remain consistent with those used in previous experiments. This reduction in variance is designed to the large gradient of the function. Intuitively, a larger gradient indicates a steeper function, making the impact of perturbations more pronounced. As a result, a smaller variance in the perturbation distribution helps improve the reliability of gradient estimation.

We compare the RMSE(Solution Gap) and RMSE(Optimality Gap). 
\begin{itemize}
    \item RMSE (Solution Gap) is defined as the RMSE of the distance between the solution point and the optimal point $(1, 1, \dots, 1)$.
    \item RMSE (Optimality Gap) is defined as the RMSE of the distance between the obtained value and the optimal value $0$.
\end{itemize}

Table \ref{tab:comparison_fun_213} presents results for function \eqref{function213} with $d=64$, based on 200 replications for each configuration. Note that {\it the actual sample pairs used in the table is $d$ times the values listed to account for the demands of this relatively high-dimensional setting}. For the Cor-CFD-GD algorithm, we adhere to the original setup. For the SPSA algorithm, the results reflect performance under optimal parameters. 
Experimental results demonstrate that the Cor-CFD algorithm consistently outperforms the SPSA algorithm across all parameter settings tested, even when SPSA is optimized with its best parameters. For example, under setting with noise level $\sigma=1$ and sample pairs of 1000, the Cor-CFD-GD algorithm achieves RMSE (Solution Gap) of 5.46 and RMSE (Optimality Gap) of 3.67, while SPSA yields a higher RMSE (Solution Gap) of 8.29 and RMSE (Optimality Gap) of 3825.04.
The superior performance of the Cor-CFD algorithm over the SPSA algorithm can be attributed to fundamental differences in their gradient estimation strategies and step size adaptation mechanisms. 

\begin{table}[t]
\renewcommand\arraystretch{1.25}
\centering
\caption{Comparison of the Cor-CFD-GD algorithm and the SPSA algorithm for function \eqref{function213}.}
\label{tab:comparison_fun_213}
\vspace{1em}
\begin{tabular}{cccccccc}
\toprule
 &  & \multicolumn{3}{c}{RMSE (Solution Gap)} & \multicolumn{3}{c}{RMSE (Optimality Gap)} \\
\cmidrule(lr){3-5} \cmidrule(lr){6-8}
$\sigma$ & Method & \multicolumn{1}{r}{1,000} & \multicolumn{1}{r}{5,000} & \multicolumn{1}{r}{10,000} & \multicolumn{1}{r}{1,000} & \multicolumn{1}{r}{5,000} & \multicolumn{1}{r}{10,000} \\
\midrule
\multirow{2}{*}{0.1} & Cor-CFD-GD & \multicolumn{1}{r}{4.37} & \multicolumn{1}{r}{3.22} & \multicolumn{1}{r}{2.81} & \multicolumn{1}{r}{0.43} & \multicolumn{1}{r}{0.16} & \multicolumn{1}{r}{0.12} \\
& SPSA       & \multicolumn{1}{r}{8.30} & \multicolumn{1}{r}{8.19} & \multicolumn{1}{r}{8.87} & \multicolumn{1}{r}{3832.61} & \multicolumn{1}{r}{4322.54} & \multicolumn{1}{r}{1535.91} \\
\multirow{2}{*}{1}   & Cor-CFD-GD & \multicolumn{1}{r}{5.46} & \multicolumn{1}{r}{4.10} & \multicolumn{1}{r}{3.55} & \multicolumn{1}{r}{3.67} & \multicolumn{1}{r}{1.54} & \multicolumn{1}{r}{1.28} \\
& SPSA       & \multicolumn{1}{r}{8.29} & \multicolumn{1}{r}{8.21} & \multicolumn{1}{r}{8.86} & \multicolumn{1}{r}{3825.04} & \multicolumn{1}{r}{4347.60} & \multicolumn{1}{r}{1111.25} \\
\multirow{2}{*}{10}  & Cor-CFD-GD & \multicolumn{1}{r}{6.37} & \multicolumn{1}{r}{5.25} & \multicolumn{1}{r}{4.67} & \multicolumn{1}{r}{25.77} & \multicolumn{1}{r}{17.33} & \multicolumn{1}{r}{16.54} \\
& SPSA       & \multicolumn{1}{r}{8.28} & \multicolumn{1}{r}{8.21} & \multicolumn{1}{r}{8.90} & \multicolumn{1}{r}{3816.15} & \multicolumn{1}{r}{4340.77} & \multicolumn{1}{r}{1126.00} \\
\bottomrule
\end{tabular}
\end{table}

Since SPSA estimates the gradient using only two samples, its gradient estimation may lack accuracy. For function \eqref{function213}, the gradient is large near the starting points and becomes smaller near the optimal point. To avoid overshooting at the beginning, a small initial step size would be necessary. However, as the algorithm approaches the optimal region, this small step size restricts efficient convergence, significantly slowing down the optimization as it nears the optimal point.

Cor-CFD estimates gradients using multiple samples, enhancing both the accuracy and stability of its gradient approximations. This mitigates the risk of extreme directional jumps, improving robustness across different variance conditions.
The Cor-CFD-GD algorithm employs an Armijo line search \eqref{eq:Armijo} to dynamically adjust the step size based on the gradient magnitude at each iteration. This approach allows the algorithm to adaptively select small step sizes when gradients are large (e.g., at the initial points), preventing overshooting and ensuring stable progression. As the gradient diminishes near the optimal point, the Cor-CFD-GD algorithm can take larger steps, accelerating convergence in low-gradient regions.
By combining the batch-based gradient estimator with Armijo line search, the Cor-CFD-GD algorithm maintains stability in early iterations and avoids the slow convergence issues encountered by SPSA near the optimal point. Unlike SPSA, which is highly sensitive to initial step size, Cor-CFD’s adaptive step size strategy is more resilient, allowing it to perform well without the need for careful tuning of initial parameters.

Notably, the SPSA algorithm's performance in the table remains unaffected by variance. In contrast, the performance of the Cor-CFD-GD algorithm declines as variance increases. This difference arises because SPSA’s slower convergence rate keeps it in regions with large function values relative to the noise for a longer period, where the influence of variance is minimal. 
Meanwhile, the Cor-CFD-GD algorithm descends more quickly into regions where function values are comparatively small relative to the noise, which decreases the accuracy of its gradient estimates and, consequently, affects optimization performance.

\begin{figure}[ht!]
\centering
      \includegraphics[width=1\linewidth]{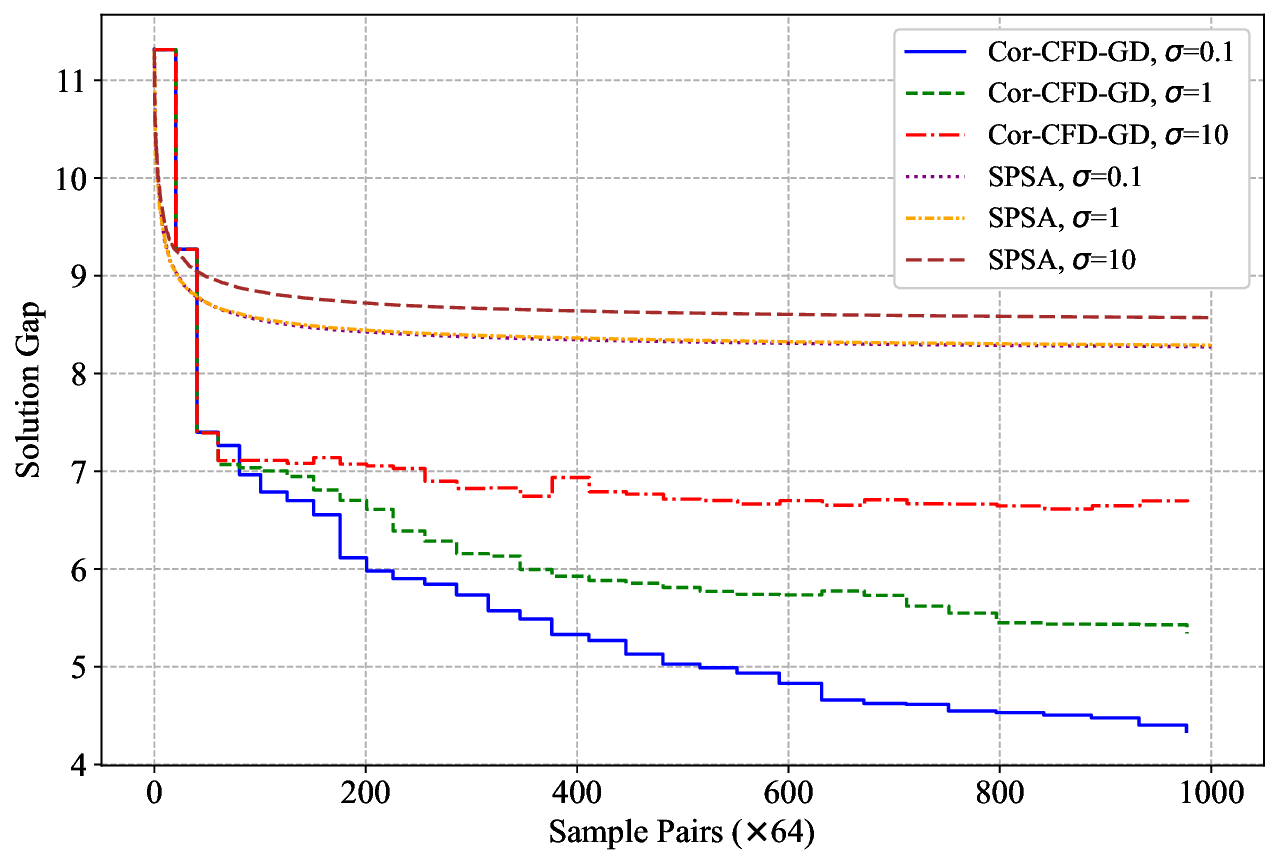}
      \caption{Solution gap comparison between Cor-CFD-GD and SPSA algorithms across sample pairs for function \eqref{function213}.}
      \label{fig:func213_point}
\end{figure}

\begin{figure}[ht!]
\centering
      \includegraphics[width=1\linewidth]{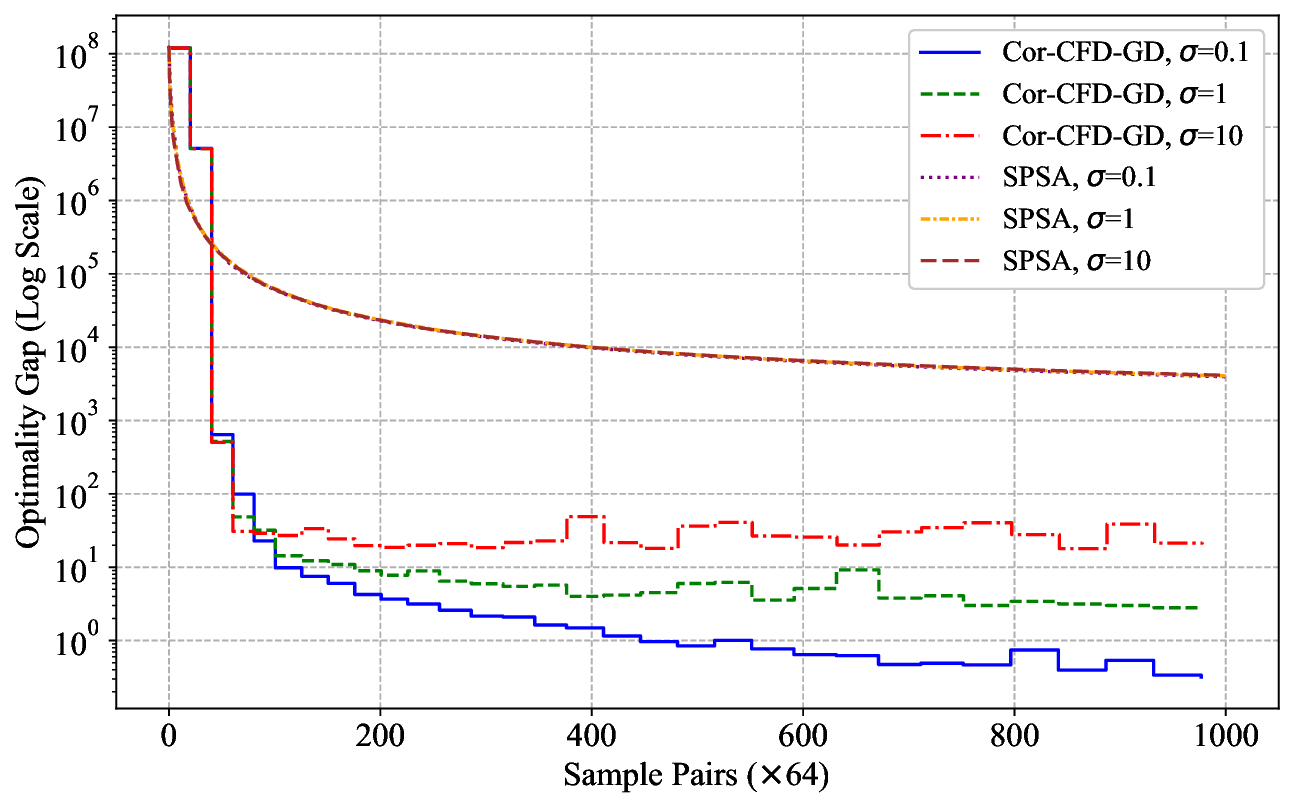}
      \caption{Optimality gap comparison between Cor-CFD-GD and SPSA algorithms across sample pairs for function \eqref{function213}.}
      \label{fig:func213_value}
\end{figure}

Fig. \ref{fig:func213_point} shows the changes in the solution gap for a single experiment with sample pairs of 1000, comparing the Cor-CFD-GD and SPSA algorithms across sample pairs. Similarly, Fig. \ref{fig:func213_value} presents the changes in the optimality gap for the Cor-CFD-GD and SPSA algorithms under the same budget, showing their respective performances over sample pairs. It is important to note that the Cor-CFD-GD algorithm does not utilize all the sample pairs, as the remaining sample pairs are insufficient for Cor-CFD-GD to conduct a new gradient estimator.
Initially, the SPSA algorithm demonstrates better performance due to its higher iteration steps within the first 2500 sample pairs. However, once the Cor-CFD-GD algorithm completes two iterations (approximately 40.36 sample pairs $\times$ 64,  where $20 \times 2$ sample pairs for Cor-CFD and 0.36 for line search), it begins to outperform the SPSA algorithm, achieving a faster and more stable reduction in the optimality gap across all noise levels. The Cor-CFD-GD’s superior long-term performance is attributed to its reliable gradient estimation method, enabling accurate descent direction and the use of larger step sizes for faster convergence. In contrast, the SPSA algorithm uses only two samples to approximate the gradient, which results in a less precise descent direction and forces the algorithm to use smaller step sizes to maintain stability.


\subsection{Hyperparameter Optimization}
In this section, we consider hyperparameter optimization, a crucial task in machine learning. 
Specifically, we address the challenge of selecting the penalty parameter $\lambda$ in ridge regression, which plays a key role in balancing model complexity and performance. We formulate the selection of the optimal penalty parameter as a DFO problem.

We use the data from \cite{house-prices-advanced-regression-techniques}, which originates from a Kaggle competition. The dataset contains 1460 samples with 81 features, and the objective is to predict house prices. 
For data preprocessing, we follow the conventional steps outlined in the Kaggle notebook \href{https://www.kaggle.com/code/apapiu/regularized-linear-models}{https://www.kaggle.com/code/apapiu/regularized-linear-models}. The preprocessing steps are as follows:
\begin{enumerate}
    \item The target variable (house price) is transformed using the natural logarithm.
    \item For numeric features, we compute the skewness of each feature and apply a log transformation (after adding 1) to those with a skewness greater than 0.75.
    \item For categorical features, we convert them to dummy variables (one-hot encoding), creating a new binary column for each possible category value to indicate its presence or absence.
    \item Missing values in the dataset are imputed using the mean of their respective columns.
\end{enumerate}
After applying these preprocessing steps, the dataset is transformed into 1460 samples with 288 features.

\begin{figure}[htbp]
      \includegraphics[width=1\linewidth]{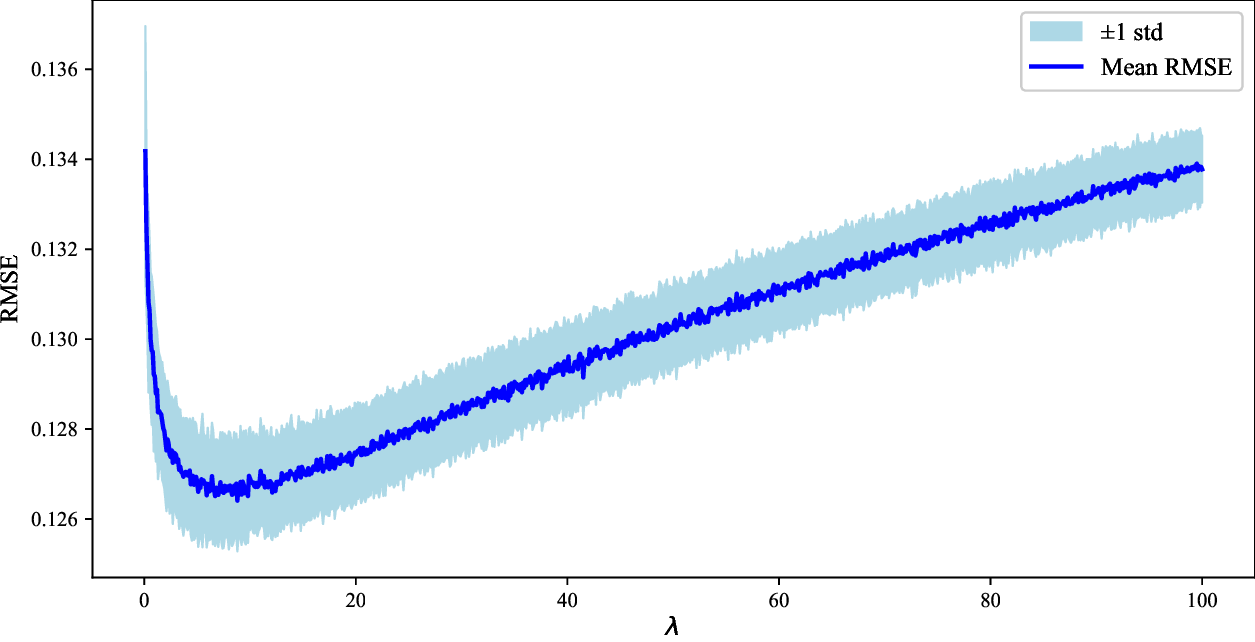}
      \caption{Effect of the regularization parameter $\lambda$ on RMSE.}
      \label{fig:HPO_function}
\end{figure}

In order to build our black-box function, we consider the algorithm presented in Algorithm \ref{alg:algorithm_HPO}.
\begin{algorithm2e}[t!]
    \caption{$K$-Fold Cross-Validation for Ridge Regression.}
    \label{alg:algorithm_HPO}
    \BlankLine
    \textit{\textbf{Input: }} Training dataset $\mathcal{D}$, number of folds $K$, penalty parameter $\lambda>0$.
    
    \textit{\textbf{Initialization: }} Randomly shuffle  $\mathcal{D}$. Split $\mathcal{D}$ into $K$ equal-sized folds: ${\mathcal{D}_1, \mathcal{D}_2, ..., \mathcal{D}_K}$.
    
    \textit{\textbf{For $k = 1$ \KwTo $K$ do}}
    \begin{enumerate}
        \item Set $\mathcal{D}_{valid} = \mathcal{D}_k$, $\mathcal{D}_{train} = \mathcal{D} \setminus \mathcal{D}_k$.
        \item Train a ridge regression model on $\mathcal{D}_{train}$ with penalty parameter $\lambda$.
        \item Use the trained model to compute the RMSE $l_k$ on $\mathcal{D}_{\text{valid}}$. 
    \end{enumerate}
    
    \textit{\textbf{Output:}} The RMSE averaged over all folds $\frac{1}{K} \sum_{k=1}^{K} l_k$.
\end{algorithm2e}
Due to the randomness in the shuffling process within the algorithm, the final results are noisy. The values returned can be regarded as outputs of a black-box function with no explicit expression. We consider the case $K=10$. By repeating 100 times at each point, sampled at intervals of 0.1 from 0.1 to 100, we can plot the sample mean and sample standard deviation (std) of this black-box function. Fig. \ref{fig:HPO_function} shows the mean and std of the RMSE.

This black-box function is extremely challenging to optimize due to its very small gradients and high noise levels. We compare the KW algorithm with the Cor-CFD-GD algorithm.

For the Cor-CFD-GD algorithm, since the ridge regression requires $\lambda>0$, it is necessary to ensure that the perturbation satisfies $x_k - c_k^r>0$ for all $r=1, 2, ..., R$. To achieve this, we set the std of the perturbation $\mathcal{P}_0$ to $\min (0.1, x_k/3)$, ensuring that most generated perturbations satisfy the condition. Any perturbations that yield values unsatisfied are regenerated. We set $a_0 = 10$. All other parameters remain consistent with our previous experiments. For the KW algorithm, we use tuning sequences $a_k=10/k$, $c_k=1/(k+20)^{1/4}$. Compared to the previous experiment, the step size has been increased to ensure a fair comparison. Additionally, a constant has been added to the denominator of the perturbation term to prevent the initial perturbation from exceeding 0.5, thus satisfying the requirements of the ridge regression.

To evaluate the performance of the solution point, we use Cubic Spline interpolation to approximate the true function mean. Considering the initial point $x_0 =0.5$, its function value is 0.1306. The optimal function value is expected to fall roughly within the range of 5 to 10, with the value around 0.1265. We report the 5th, 50th, and 95th percentiles of the solution point, denoted by ``5\%'', ``Median'', and ``95\%'', respectively. We also report the sample mean and std of the estimated values under different budgets. Table \ref{tab:comparison_HPO} presents the results, each based on 100 replications.
\begin{table}[t]
\renewcommand\arraystretch{1.25}
    \centering
    \caption{Comparison of the Cor-CFD-GD algorithm and the KW algorithm for hyperparameter optimization.}
    \label{tab:comparison_HPO}
    \begin{tabular}{c c c c c c c}
        \toprule
         &  & \multicolumn{3}{c}{Quantile of solution point} & \multicolumn{2}{c}{Estimated Value} \\
        \cmidrule(lr){3-5} \cmidrule(lr){6-7}
        Sample pairs & Method & 5\% & Median & 95\% & Mean & Std \\
        \midrule
        \multirow{2}{*}{1,000} & Cor-CFD-GD & \multicolumn{1}{r}{0.7439} & \multicolumn{1}{r}{1.3094} & \multicolumn{1}{r}{2.4364} & \multicolumn{1}{r}{0.1286} & \multicolumn{1}{r}{0.00096} \\
         & KW & \multicolumn{1}{r}{0.7790} & \multicolumn{1}{r}{0.8271} & \multicolumn{1}{r}{0.8803} & \multicolumn{1}{r}{0.1296} & \multicolumn{1}{r}{0.00015} \\
        \multirow{2}{*}{5,000} & Cor-CFD-GD & \multicolumn{1}{r}{1.3066} & \multicolumn{1}{r}{2.2526} & \multicolumn{1}{r}{15.6276} & \multicolumn{1}{r}{0.1278} & \multicolumn{1}{r}{0.00073} \\
         & KW & \multicolumn{1}{r}{0.8049} & \multicolumn{1}{r}{0.8437} & \multicolumn{1}{r}{0.8854} & \multicolumn{1}{r}{0.1295} & \multicolumn{1}{r}{0.00012} \\
        \multirow{2}{*}{10,000} & Cor-CFD-GD & \multicolumn{1}{r}{1.1991} & \multicolumn{1}{r}{2.6045} & \multicolumn{1}{r}{22.9126} & \multicolumn{1}{r}{0.1276} & \multicolumn{1}{r}{0.00061} \\
         & KW & \multicolumn{1}{r}{0.8249} & \multicolumn{1}{r}{0.8606} & \multicolumn{1}{r}{0.8966} & \multicolumn{1}{r}{0.1294} & \multicolumn{1}{r}{0.00011} \\
        \bottomrule
    \end{tabular}
    \label{tab:budget_quantiles}
\end{table}

With 1000 sample pairs, the Cor-CFD-GD algorithm reduces the mean of the estimated value from an initial 0.1306 to 0.1286, while the KW algorithm results in a higher mean of 0.1296. Notably, the KW algorithm exhibits a lower std (0.00015 compared to 0.00096 for the Cor-CFD-GD algorithm), indicating stability. This stability can be largely attributed to the decaying step size used in the KW algorithm, which effectively prevents large jumps during iterations. This trend can be observed across various sample pairs cases. However, despite its stability, at 5000 sample pairs, the mean of the estimated values of the KW algorithm exceeds that of the Cor-CFD-GD algorithm by more than twice the std of the Cor-CFD-GD algorithm, and at 10000 sample pairs, this difference reaches three times the std, illustrating the improved performance of the Cor-CFD-GD algorithm. 

For the Cor-CFD-GD algorithm, the quantile data reveals that the distribution of the solution point is gradually toward the true optimal range of 5 to 10. For example, at 1000 sample pairs, the median is 1.3094, which is well below the optimal range. While at 5000 sample pairs, the median increases to 2.2526. It is important to note that since the function is U-shaped, a larger solution point does not necessarily correspond to a higher estimated value. Although the KW algorithm yields a much narrower range, with 10000 sample pairs, the 95\% quantile of the Cor-CFD-GD algorithm is 22.91 (with an estimated value of 0.1278), while the 95\% quantile of the KW algorithm is 0.8966 (with an estimated value of 0.1292), indicating that the Cor-CFD algorithm yields better performance.

Furthermore, for the Cor-CFD-GD algorithm, the largest final solution point is 33.66 in sample pairs 10000 with an estimated value of 0.1288, which is still lower than the mean of the KW algorithm. By analyzing the changes during the iterations, we find that in the early stages, the small sample size $n_0$ leads to an imprecise estimation of the gradient direction, causing the iteration points to move closer to zero. However, when the iteration points are near zero, the perturbation should be very small to ensure that $x_k - c_k^r>0$ holds. This, in turn, results in inaccurate gradient estimation, leading to a sudden jump to a distant location within a single iteration. A potential improvement is to increase the initial sample size or use a one-sided finite difference method for gradient estimation.

\section{Conclusions}\label{section4}

In this paper, we have conducted experimental studies to investigate the trade-off between gradient estimation accuracy and iteration steps in DFO. 
Our results demonstrate that the Cor-CFD method, which prioritizes gradient estimation accuracy through careful sample utilization, outperforms traditional approaches such as KW and SPSA, which favor frequent iterations with minimal samples.
This superior performance is observed across both low- and high-dimensional settings, as well as a hyperparameter optimization problem, suggesting that the benefits of accurate gradient estimation outweigh the computational cost of additional samples per iteration.

Several promising directions for future research emerge from this work. 
First, developing adaptive sample allocation strategies that dynamically adjust the number of samples based on the optimization results could further enhance the efficiency of gradient estimation. \cite{bollapragada_progressive_2018} presents a promising solution for the L-BFGS algorithm.
The second direction involves establishing the theoretical convergence properties of optimization algorithms based on Cor-CFD. The theory presented in \cite{hu2024convergence} may offer some valuable guidance for this direction.
Finally, exploring the application of these findings to specific domains, such as deep learning hyperparameter optimization and reinforcement learning, could yield practical benefits in these increasingly important fields.

\bibliographystyle{plainnat}
\bibliography{mybibfile}

\end{document}